\documentclass[11pt]{article}

\usepackage[final]{acl}

\usepackage[dvipsnames]{xcolor}
\usepackage{times}
\usepackage{latexsym}
\usepackage[T1]{fontenc}
\usepackage[utf8]{inputenc}

\usepackage{enumitem}
\setlist[itemize]{noitemsep}

\usepackage{microtype}
\usepackage{inconsolata}
\usepackage{graphicx}
\usepackage{subcaption}
\usepackage{multirow}
\usepackage{amssymb}
\usepackage{tabularx}

\title{HalluMix: A Task-Agnostic, Multi-Domain Benchmark for Real-World Hallucination Detection}

\author{
 \textbf{Deanna Emery}, \\
 \textbf{Michael Goitia}, \textbf{Freddie Vargus}, \textbf{Iulia Neagu}
\\
  Quotient AI
\\
 \small{
   \texttt{\{deanna, mike, freddie, julia\}@quotientai.co}
 }
}

\begin{document}

\maketitle

\begin{abstract}
As large language models (LLMs) are increasingly deployed in high-stakes domains, detecting hallucinated content—text that is not grounded in supporting evidence—has become a critical challenge. Existing benchmarks for hallucination detection are often synthetically generated, narrowly focused on extractive question answering, and fail to capture the complexity of real-world scenarios involving multi-document contexts and full-sentence outputs. 
We introduce the \textit{HalluMix} Benchmark, a diverse, task-agnostic dataset that includes examples from a range of domains and formats. Using this benchmark, we evaluate seven hallucination detection systems—both open and closed source—highlighting differences in performance across tasks, document lengths, and input representations. Our analysis highlights substantial performance disparities between short and long contexts, with critical implications for real-world Retrieval Augmented Generation (RAG) implementations. \textit{Quotient Detections} achieves the best overall performance, with an accuracy of 0.82 and an F1 score of 0.84.
\end{abstract}

\section{Introduction}\label{section:Introduction}

As large language models (LLMs) continue to gain prominence across domains, ensuring the factual correctness of their outputs has become a central concern. A critical issue in this context is \textit{hallucination}, where the model generates content not supported by, or contradictory to, a given source. In high-stakes fields such as law, medicine, and finance, such hallucinations can undermine trust and lead to harmful consequences \cite{hallucinations-survey}.

While detecting hallucinations remains an active area of research, progress has been hindered by the lack of representative benchmarks. Most existing evaluation datasets are task-specific—often focused on open-book question answering—and rely heavily on synthetic examples or narrow context formats \cite{hallucinations-survey, lynx, halueval, ragtruth, hotpotqa}. This limits their generalizability to real-world settings, where LLM outputs are typically multi-sentence or paragraph responses grounded in multi-document contexts.

We introduce the \textbf{HalluMix} Benchmark, a large-scale, domain-diverse dataset specifically designed to evaluate hallucination detection in realistic generation scenarios. Our dataset includes examples drawn from multiple tasks—including summarization, question answering, and natural language inference—and spans a wide array of domains such as healthcare, law, science, and news. Each instance consists of a multi-document context and a response, with binary hallucination labels indicating whether the response is faithful to the provided documents.

We further use this benchmark to systematically evaluate seven state-of-the-art hallucination detection systems, including both open source and commercial tools.

Our contributions are threefold: 
\begin{itemize}
    \item We propose a unified benchmark for hallucination detection, constructed from high-quality human-curated datasets spanning multiple tasks and domains.
    \item We introduce a consistent evaluation framework that decouples hallucination detection from task-specific assumptions (e.g. the presence of a question), reflecting more diverse LLM use-cases.
    \item We conduct a comparative evaluation of existing hallucination detection methods, providing insights into their strengths, weaknesses, and suitability for different real-world applications.
\end{itemize}

The following sections detail our benchmark construction methodology, present our comparative evaluation of leading hallucination detection systems, and discuss the implications of our findings for both academic and industry applications.

\section{The HalluMix Benchmark}\label{section:Benchmark-Dataset}

Existing hallucination benchmark datasets are often generated by LLMs and are heavily focused on question-answering tasks. In many cases, the reference answers in these datasets are limited to single-word spans extracted directly from a single context, limiting their applicability to more complex forms of generation. \cite{lynx, halueval, ragtruth, hotpotqa}.

However, hallucinations are not confined to question-answering alone; they frequently arise in other tasks such as summarization, dialogue, and open-ended generation. Furthermore, real-world LLM deployments typically use \textit{lists} of documents, often extracted via Retrieval Augmented Generation (RAG), to produce \textit{full-sentence} outputs rather than short extractive spans. To address these limitations, we constructed a new benchmark that decouples hallucination detection from the question-answering format. Each example consists of a context (split into a list of text segments) and a response, enabling evaluation based solely on the factual consistency between the two.

To evaluate the performance of hallucination detectors, we constructed the \textit{HalluMix} Benchmark, a diverse dataset that integrates samples from multiple human-curated sources. These include tasks from summarization, natural language inference (NLI), and question-answering (QA), covering a wide array of domains such as news, science, law, healthcare, dialogues, and long-form narratives. Each example is labeled as either \textit{faithful} or \textit{hallucinated}.

\begin{figure}[t]
    \centering
    \includegraphics[width=1\linewidth]{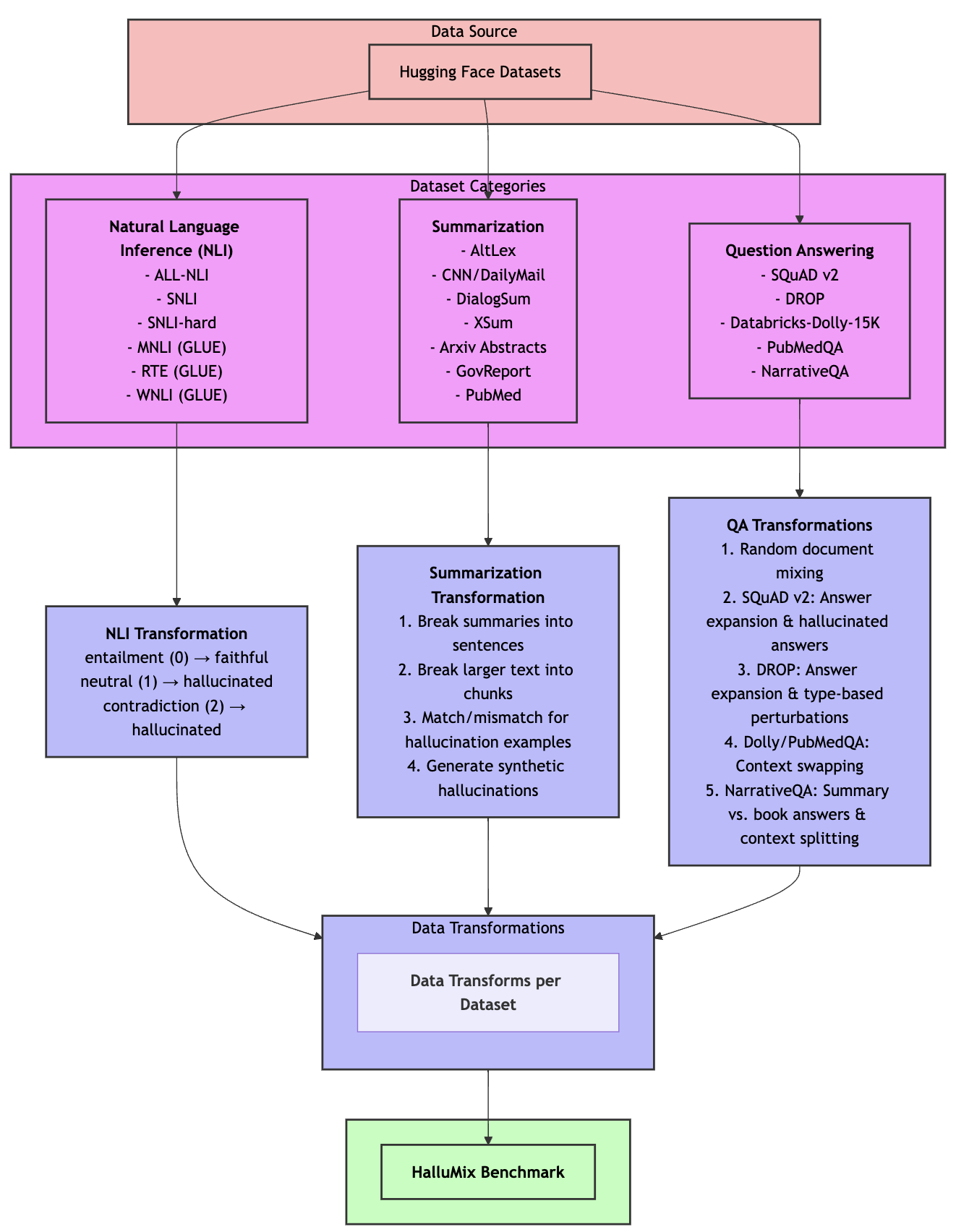}
    \caption{Overview of the HalluMix construction pipeline, showing datasets and transformation strategies.}
    \label{fig:benchmark-diagram}
\end{figure}

We selected datasets that are predominantly human-labeled or human-curated. As shown in the data process diagram in Figure~\ref{fig:benchmark-diagram}, we applied a variety of dataset-specific transformations to construct hallucinated examples, while using original annotations for faithful samples wherever possible.

\subsection{Data Transformations}

\subsubsection{Natural Language Inference Datasets}

NLI datasets were repurposed for hallucination detection by reinterpreting their label schema. Each example consists of a \textit{premise} and a \textit{hypothesis}, along with a label indicating their relationship. We used the following mapping to convert the NLI labels to hallucination labels:

\begin{itemize}
    \item \textit{Faithful:} Hypotheses labeled as \texttt{entailment}.
    \item \textit{Hallucinated:} Hypotheses labeled as \texttt{neutral} or \texttt{contradiction}.
\end{itemize}

In datasets with binary NLI labels (\texttt{entailment} vs. \texttt{non-entailment}), we applied a similar mapping, treating \texttt{non-entailment} as hallucinated. The following NLI datasets were used:
\begin{itemize}
    \item \texttt{sentence-transformers/all-nli} \cite{multi-nli, snli}
    \item \texttt{stanfordnlp/snli} \cite{snli}
    \item \texttt{snli-hard} \cite{snli-hard}
    \item \texttt{glue: mnli, rte, and wnli} \cite{glue}
\end{itemize}

\subsubsection{Summarization Datasets}

Summarization datasets consist of long-form documents paired with human-written summaries. Since summaries are designed to be faithful to the original documents, we label these as \textit{faithful} by default. To create hallucinated examples, we apply a permutation-based transformation: summaries are randomly mismatched with unrelated documents.

We include the following summarization datasets:
\begin{itemize}
    \item \texttt{sentence-transformers/altex} \cite{altex}
    \item \texttt{CNN/DailyMail} \cite{cnn-dailymail}
    \item \texttt{DialogSum} \cite{dialogsum}
    \item \texttt{XSum} \cite{xsum}
    \item \texttt{arXiv summarization} \cite{arxiv-summarization}
    \item \texttt{GovReport summarization} \cite{govreport-summarization}
    \item \texttt{PubMed summarization} \cite{pubmed-summarization}
\end{itemize}

\subsection{Question Answering Datasets}

QA datasets contain a question, a context passage, and a corresponding answer. By design, the answers in these datasets are \textit{faithful}. Hallucinated variants are generated via multiple, dataset-specific strategies. In some datasets, answers consisted of single words; we used an LLM to expand these into complete declarative sentences (e.g., \textit{Q: What color is the car? A: Red} $\rightarrow$ \textit{The car is red.}) to ensure alignment with real-world use cases and to separate dependence on questions to determine hallucinations.

We used the following QA datasets:
\begin{itemize}
    \item \texttt{SQuAD-v2} \cite{squad,squadv2}: Unanswerable questions (with blank answers) were paired with LLM-generated answers based solely on the question (without context), labeled as hallucinated.

    \item \texttt{DROP} \cite{drop}: Each context includes multiple questions with typed answers (numeric, date, string). Hallucinated examples were created by replacing the correct answer with another plausible answer of the same type from within the same context.

    \item \texttt{Databricks-Dolly-15K} \cite{databricks-dolly-15k} and \texttt{PubMedQA} \cite{pubmedqa}: Hallucinated examples were generated by mismatching answers and contexts.

    \item \texttt{NarrativeQA} \cite{narrativeqa}: This dataset contains book-length texts, summaries, questions, and answers. For tractability, we primarily used the document summaries as contexts. To preserve long-context evaluation, we retained a small sample of shorter full texts. Hallucinated examples were generated by mismatching answers with unrelated summaries or passages.
\end{itemize}

\subsection{Final Dataset Structure}

HalluMix is structured to support robust and flexible hallucination detection evaluation. 

To better reflect real-world information retrieval scenarios (i.e., RAG), each context was split into even-sized chunks consisting of complete sentences, ensuring that no chunk contained partial or fragmented sentences. This approach preserved grammatical integrity while maintaining manageable chunk lengths and preventing over-segmentation. Additionally, we randomly shuffled the document chunks to remove any ordering advantages, as real-world retrieval systems often return documents without preserving the original narrative sequence.

To simulate realistic retrieval noise, we augmented faithful examples with ten randomly selected, irrelevant document chunks from unrelated documents within the benchmark. The added content increases the challenge of identifying relevant information without altering the evidence available for grounding the hypothesis. By applying this augmentation exclusively to faithful examples, we avoided inadvertently introducing supporting evidence into hallucinated cases. This approach creates an evaluation environment that mirrors real-world conditions where hallucination detection systems must succeed despite noisy document retrieval.

Each example in the final dataset includes:
\begin{itemize}
    \item A \textbf{documents} field: the context represented as a list of text chunks (e.g., tokenized sentences or paragraph blocks),
    \item An \textbf{answer}: the hypothesis to be evaluated, such as a summary sentence, answer, or claim,
    \item A binary \textbf{hallucination label}: where \texttt{0} denotes \textit{faithful} and \texttt{1} denotes \textit{hallucinated},
    \item A \textbf{source identifier}: to indicate the original dataset for provenance tracking.
\end{itemize}
Representative examples of hallucinated and faithful data points are provided in Tables \ref{tab:data-example-hallucinated} and \ref{tab:data-example-faithful} in the Appendix.

Faithful examples (\texttt{label = 0}) come directly from human-labeled or human-curated datasets. Hallucinated examples (\texttt{label = 1}) were constructed, in some cases, through controlled transformations such as summary mismatches, QA context permutations, or NLI relabeling. Due to these transformations—including chunking, shuffling, distractor insertion, and label reassignment—each data point in HalluMix has been substantially modified and should not be considered equivalent to its original source, even when the source identifier is preserved for tracking purposes.

The final dataset was de-duplicated and a stratified random sample of 6.5k data points was collected to achieve a balanced dataset with equal representation across hallucination labels, data types and sources. Each source has roughly equal representation within each data type (NLI, QA, Summarization), and each data type has roughly equal representation across the benchmark dataset.

The resulting dataset offers a unified and extensible benchmark for hallucination detection across multiple domains, formats, and task settings.

\section{Benchmarking Methodology}\label{section:Benchmarking-Methodology}

Using the \textit{HalluMix} Benchmark, we compared the performance of different hallucination detectors, selecting methods based on practical deployment considerations including model size ($\le$8B parameters), inference cost, and latency requirements. Our evaluation includes both open source and closed source approaches:

\begin{itemize}
    \item \textit{Llama-3-Patronus-Lynx-8B-Instruct-v1.1} \cite{lynx} - A Llama-3.1 model, fine-tuned on hallucination detection datasets. The output is a binary score.
    \item \textit{Ragas Faithfulness} \cite{ragas} - A two-step approach that uses an LLM to identify the distinct claims within the model response, then uses an LLM-as-a-Judge to determine whether each claim is faithful to the source documents. The output is the fraction of claims that are faithful to the documents (i.e. a value less than 1 indicates presence of hallucination).
    \item \textit{Azure Groundedness} \cite{azure-groundedness} - A closed source API-based hallucinations detector. The output is a binary score.
    \item \textit{Vectara HHEM-2.1-Open} \cite{hhem-2.1-open} - An open-weights version of the HHEM-2.1 model. The output is a likelihood (between 0-1) of faithfulness. In this paper, we set a threshold such that values less than 0.5 are predicted as \textit{hallucinated}.
    \item \textit{Vertex AI Grounding} \cite{vertex-grounding} - A closed source API-based hallucinations detector. The output is a likelihood (between 0-1) of faithfulness. In this paper, we set a threshold such that values less than 0.5 are predicted as \textit{hallucinated}.
    \item \textit{Bespoke-Minicheck-7B }\cite{minicheck, bespokeminicheck} - A fine-tuned 7B parameter model that accepts a sentence and document pair for evaluation. Multi-sentence responses are first tokenized into sentences before evaluation. The model returns a binary score for each sentence. In this paper, if any sentence is predicted to be hallucination, we set the overall prediction to \textit{hallucinated}.
    \item \textit{Quotient Detections} - An LLM-as-a-Judge that uses a sentence-based approach to identify hallucinations. The output is a binary score indicating that at least one sentence contains a hallucination.
\end{itemize}

\begin{table*}[t]
    \centering
    \begin{tabular}{lcccc}
         & Single Context & List of Documents & Question & Response \\
        \hline
        Quotient Detections & - & \checkmark & - & \checkmark \\
        Patronus Lynx 8B & \checkmark & - & \checkmark & \checkmark \\
        Ragas Faithfulness & - & \checkmark & \checkmark & \checkmark \\
        Azure Groundedness & - & \checkmark & \textit{Optional} & \checkmark \\
        Vectara HHEM-2.1-Open & \checkmark & - & - & \checkmark \\
        Vertex AI Grounding & - & \checkmark & - & \checkmark \\
        Bespoke-Minicheck-7B & - & \checkmark & - & \checkmark \\
        \hline
    \end{tabular}
    \caption{Input requirements and formats for each hallucination detection method. A checkmark indicates that the field is required. A dash indicates that the field is not accepted. For Azure Groundedness, there are separate API request formats for QA and summarization tasks, enabling hallucination detection both with and without a question.}
    \label{table:method-inputs}
\end{table*}

The required inputs and input formats for each of these hallucination detectors are different. Some require a single context, while others accept a list of documents; some require a question as an input, while others only need context and response. Table \ref{table:method-inputs} lists the input requirements for each hallucination detection method. 

Because our benchmark dataset is question-agnostic, not all examples have an applicable question; in these cases, we input the question as \texttt{None} when required by the detector. For instances where the detector did not accept a list of documents, we instead join the documents with two new-line separators between each document. Both Patronus Lynx 8B and Vectara HHEM-2.1-Open required a single context input.

\section{Results}\label{section:Results}

\begin{table*}[t]
    \centering
    \begin{tabular}{lcccc}
         & Accuracy & F1 & Precision & Recall \\
        \hline
        Quotient Detections & \textbf{0.821} & \textbf{0.840} & 0.764 & 0.932 \\
        Bespoke Minicheck 7B & 0.808 & 0.832 & 0.744 & 0.944 \\
        Patronus Lynx 8B & 0.808 & 0.828 & 0.754 & 0.919 \\
        Ragas Faithfulness & 0.787 & 0.818 & 0.719 & \textbf{0.950} \\
        Azure Groundedness* & 0.784 & 0.788 & \textbf{0.781} & 0.795 \\
        Vectara HHEM-2.1-Open & 0.749 & 0.771 & 0.715 & 0.836 \\
        Vertex AI Grounding & 0.727 & 0.772 & 0.668 & 0.915 \\
        \hline
    \end{tabular}
    \caption{Hallucination detection performance across methods evaluated on the full benchmark dataset. Quotient Detections achieves the highest overall accuracy and F1 score, demonstrating balanced precision and recall. Azure Groundedness\textsuperscript{\ref{azure-perf-note}} attains the highest precision but with low recall, whereas Ragas Faithfulness achieves the highest recall at the expense of precision.}
    \label{table:performance}
\end{table*}

We evaluated seven hallucination detection systems on the HalluMix Benchmark, with performance metrics shown in Table~\ref{table:performance}. 

\textit{Quotient Detections} achieved the highest overall performance, leading in both Accuracy (0.82) and F1 score (0.84), while maintaining a strong balance between Precision (0.76) and Recall (0.93). While Quotient Detections didn't achieve the highest individual precision or recall scores, the methods that did excel in one metric showed substantial decline in the other. For instance, \textit{Azure Groundedness}\footnote{Azure Groundedness performance may be overestimated due to its limits on the length of input documents. We were unable to get Azure Groundedness evaluations on 304 of the longest context examples. Generally, these long context examples are more challenging.\label{azure-perf-note}} demonstrates high precision (0.78) but achieves lower recall (0.79). Conversely, \textit{Ragas Faithfulness} shows high recall (0.95) but at the cost of precision (0.72).

\begin{table*}[t]
    \centering
    \resizebox{\textwidth}{!}{
    \begin{tabular}{llccccccc}
         &  & Quotient & Bespoke & Patronus & Ragas & Azure & Vectara & Vertex AI \\
        Data Type & Data Source & Detections & Minicheck 7B & Lynx 8B & Faithfulness & Groundedness\textsuperscript{\ref{azure-perf-note}} & HHEM-2.1-Open & Grounding \\
        \hline
        \multirow{7}{*}{NLI} & au123/snli-hard & 0.856 & \textbf{0.921} & \textcolor{Mahogany}{0.686} & 0.813 & 0.878 & 0.789 & 0.694 \\
         & nyu-mll/glue/mnli & 0.847 & 0.847 & \textcolor{Mahogany}{0.662} & 0.759 & \textbf{0.912} & 0.802 & 0.751 \\
         & nyu-mll/glue/rte & 0.900 & \textbf{0.966} & 0.746 & 0.870 & 0.852 & \textcolor{Mahogany}{0.678} & 0.818 \\
         & nyu-mll/glue/wnli & 0.850 & \textbf{0.902} & 0.758 & 0.766 & 0.568 & \textcolor{Mahogany}{0.488} & 0.502 \\
         & sentence-transformers/all-nli & 0.901 & 0.908 & \textcolor{Mahogany}{0.673} & 0.821 & \textbf{0.936} & 0.821 & 0.755 \\
         & stanfordnlp/snli & \textbf{0.885} & 0.880 & \textcolor{Mahogany}{0.699} & 0.858 & 0.869 & 0.833 & 0.749 \\
        \hline
        \multirow{5}{*}{\parbox{1.8cm}{Question Answering}} & PubMedQA & 0.624 & 0.572 & \textbf{0.928} & 0.586 & 0.596 & 0.670 & \textcolor{Mahogany}{0.542} \\
         & databricks-dolly-15k & \textbf{0.880} & \textcolor{Mahogany}{0.766} & 0.826 & 0.848 & 0.842 & 0.864 & 0.855 \\
         & DROP & 0.806 & 0.766 & \textbf{0.878} & 0.736 & 0.708 & \textcolor{Mahogany}{0.478} & 0.586 \\
         & narrativeqa & 0.886 & 0.858 & \textbf{0.916} & 0.864 & \textcolor{Mahogany}{0.748} & 0.832 & 0.758 \\
         & squad\_v2 & \textbf{0.920} & 0.912 & 0.890 & 0.892 & 0.912 & \textcolor{Mahogany}{0.818} & 0.890 \\
        \hline
        \multirow{6}{*}{Summarization} & sentence-transformers/altlex & 0.883 & 0.838 & \textcolor{Mahogany}{0.730} & 0.820 & \textbf{0.932} & 0.865 & 0.869 \\
         & arxiv\_summarization & 0.614 & 0.591 & \textbf{0.926} & 0.702 & \textcolor{Mahogany}{0.568} & \textbf{0.926} & 0.633 \\
         & cnn\_dailymail & \textcolor{Mahogany}{0.753} & 0.813 & 0.822 & 0.808 & 0.817 & 0.881 & \textbf{0.936} \\
         & dialogsum & 0.876 & 0.814 & 0.814 & 0.850 & \textbf{0.920} & 0.690 & \textcolor{Mahogany}{0.606} \\
         & govreport\_summarization & 0.597 & 0.509 & \textbf{0.943} & 0.703 & \textcolor{Mahogany}{0.500} & 0.915 & 0.882 \\
         & pubmed\_summarization & 0.629 & \textcolor{Mahogany}{0.582} & \textbf{0.911} & 0.695 & 0.615 & 0.892 & 0.803 \\
         & xsum & 0.715 & 0.720 & 0.725 & 0.617 & \textbf{0.798} & 0.606 & \textcolor{Mahogany}{0.601} \\
        \hline                                                       
    \end{tabular}}
    \caption{Accuracy scores of hallucination detection methods across data sources within the benchmark. \textbf{Bold} values indicate the highest accuracy for each data source, while \textcolor{Mahogany}{red} values indicate the lowest. The substantial variation across datasets suggests method-specific strengths, with some detectors excelling on specific data types while underperforming on others—highlighting potential specialization or overfitting concerns in current hallucination detection approaches.}    
    \label{table:performance-by-source}
\end{table*}

When examining performance across different data sources (Table~\ref{table:performance-by-source}), we observe substantial variance both between methods and across datasets. 

The most striking pattern emerges in Summarization tasks, where performance diverges dramatically. \textit{Patronus Lynx 8B} consistently outperforms other approaches on long-form summarization tasks. For example, on PubMed summarization, it achieves 0.91 accuracy, compared to 0.63 for \textit{Quotient Detections} and 0.58 for \textit{Bespoke-Minicheck-7B}.

Table \ref{table:performance-by-source} shows the accuracy scores of the hallucination detection methods on each of the sub-data sources within HalluMix. The high variance in scores both across methods and across datasets indicates that each detection method likely has different strengths and weaknesses. Of note, the methods that perform generally best in the NLI and Question-Answering subsets tend to perform more poorly on the Summarization subsets and vice versa.

\begin{table}[]
    \centering
    \begin{tabular}{lcc}
         & Avg Response & Avg Document \\
        Data Type & Token Count & Token Count \\
        \hline
        NLI & 11 & 88 \\
        QA & 32 & 167 \\
        Summ. & 174 & 439 \\
        \hline
    \end{tabular}
    \caption{Average token counts for each data type (NLI, Question-Answering, and Summarization) in HalluMix.}
    \label{table:average-token-counts}
\end{table}

Table~\ref{table:average-token-counts} illustrates the substantial differences in content length across data types. NLI examples are concise (averaging 11 tokens for responses and 88 for documents), while Summarization examples involve much longer text (averaging 174 tokens for responses and 439 for documents). These length differences correlate strongly with detection performance, suggesting that the quality hallucination detection methods is dependent on the content length.

\begin{figure*}[t!]
    \centering
    \begin{subfigure}{0.451\textwidth}
        \centering
        \includegraphics[width=\linewidth]{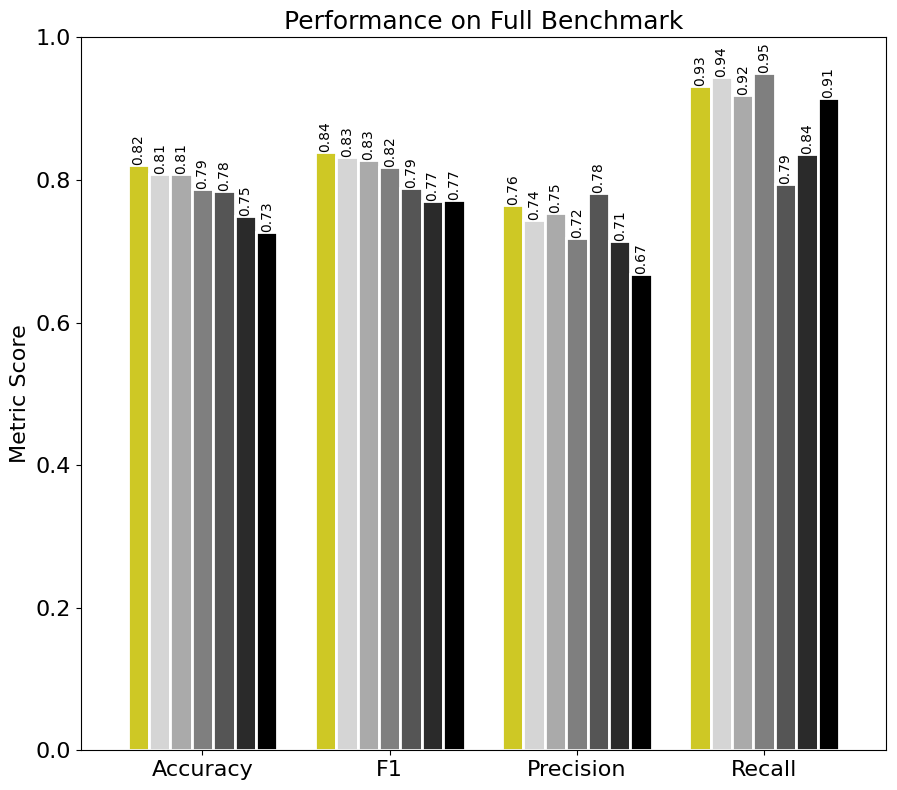}
        \caption{Performance on the full HalluMix benchmark \\ (including summarization data).}
        \label{fig:with-sum}
    \end{subfigure}%
    ~
    \begin{subfigure}{0.549\textwidth}
        \centering
        \includegraphics[width=\linewidth]{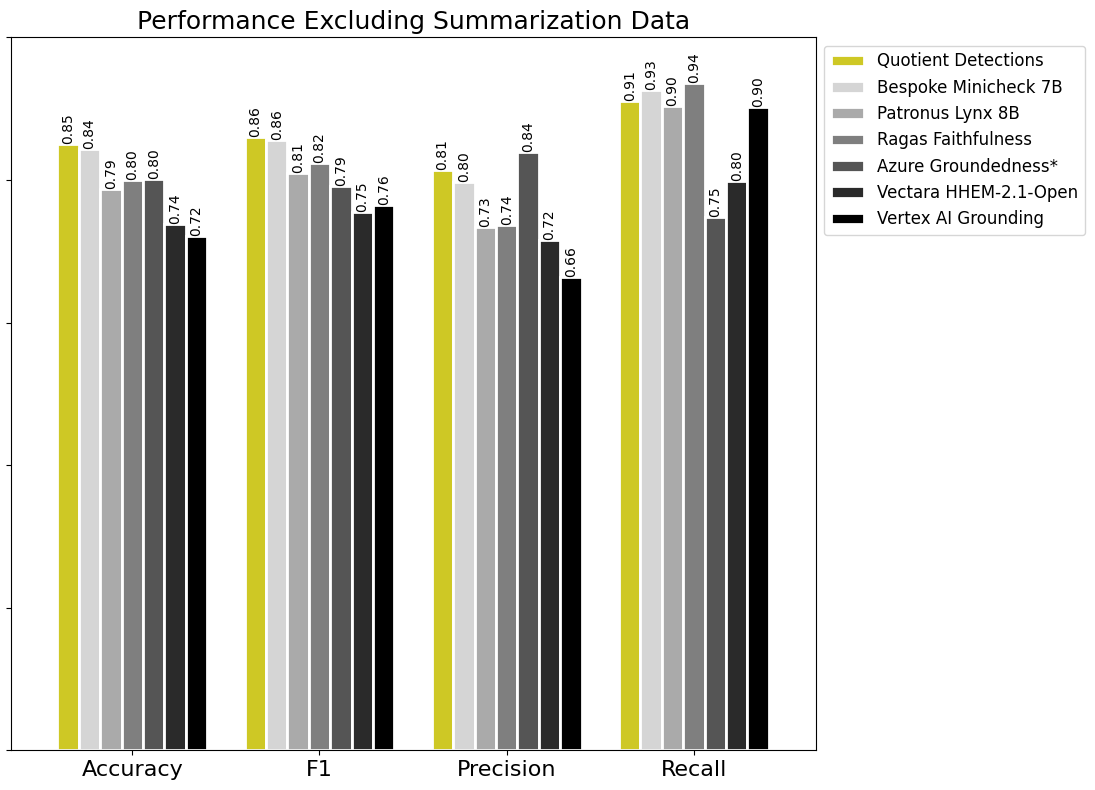}
        \caption{Performance on the HalluMix benchmark \\ excluding summarization data.}
        \label{fig:without-sum}
    \end{subfigure}
    \caption{Comparison of hallucination detection performance metrics (Accuracy, F1, Precision, Recall) across all evaluated methods. Panel (a) shows performance on the complete benchmark dataset, while panel (b) shows performance excluding summarization examples. \textit{Quotient Detections} achieves highest accuracy and F1 in both scenarios.}
    \label{fig:performance-bar-plot}
\end{figure*}

Figure~\ref{fig:performance-bar-plot} further demonstrates how content type affects relative performance. When evaluating only on shorter-context examples (NLI and QA subsets, panel b), \textit{Patronus Lynx 8B} drops from third place to fifth in accuracy, while \textit{Quotient Detections} maintains its lead. This shift underscores how benchmark composition significantly influences performance rankings.

\section{Discussion}

Our comprehensive evaluation reveals several key insights about the current state of hallucination detection systems. While the best-performing models achieve respectable accuracy overall, their effectiveness varies depending on task type, content length, and input format. These variations reflect both the diversity of our benchmark dataset and the design decisions embedded in each detection method.

\subsection{Evidence of Sub-Source Overfitting}

Table \ref{table:performance-by-source} shows that some detection systems perform exceptionally well on specific datasets while underperforming on others. This pattern suggests that certain hallucination detection methods may have been trained on or heavily influenced by particular sub-datasets, especially within the NLI and QA categories. For instance, high accuracy on well-known datasets such as SNLI or SQuAD could indicate exposure during pretraining or fine-tuning.  While this may not invalidate the performance, it does raise questions about generalizability—particularly in less conventional or domain-specific generation scenarios.

An intriguing finding from our analysis is that specialized, fine-tuned models like \textit{Patronus Lynx 8B}, \textit{Vectara HHEM-2.1-Open}, and \textit{Bespoke-Minicheck-7B}—despite being explicitly optimized for hallucination detection—do not generally outperform methods that leverage general-purpose language models with appropriate prompting strategies, such as \textit{Ragas Faithfulness} and \textit{Quotient Detections}.

\subsection{Content Length and Context Representation Challenges}

As shown in Table~\ref{table:average-token-counts}, the Summarization subset involves substantially longer contexts and responses than NLI or QA. The performance drop across most models on summarization examples suggests that long-form generation introduces additional challenges for hallucination detection, such as tracking referents, maintaining discourse coherence, and grounding claims across large textual spans.

\begin{figure*}[t!]
    \centering
    \begin{subfigure}{0.5\textwidth}
        \centering
        \includegraphics[width=\linewidth]{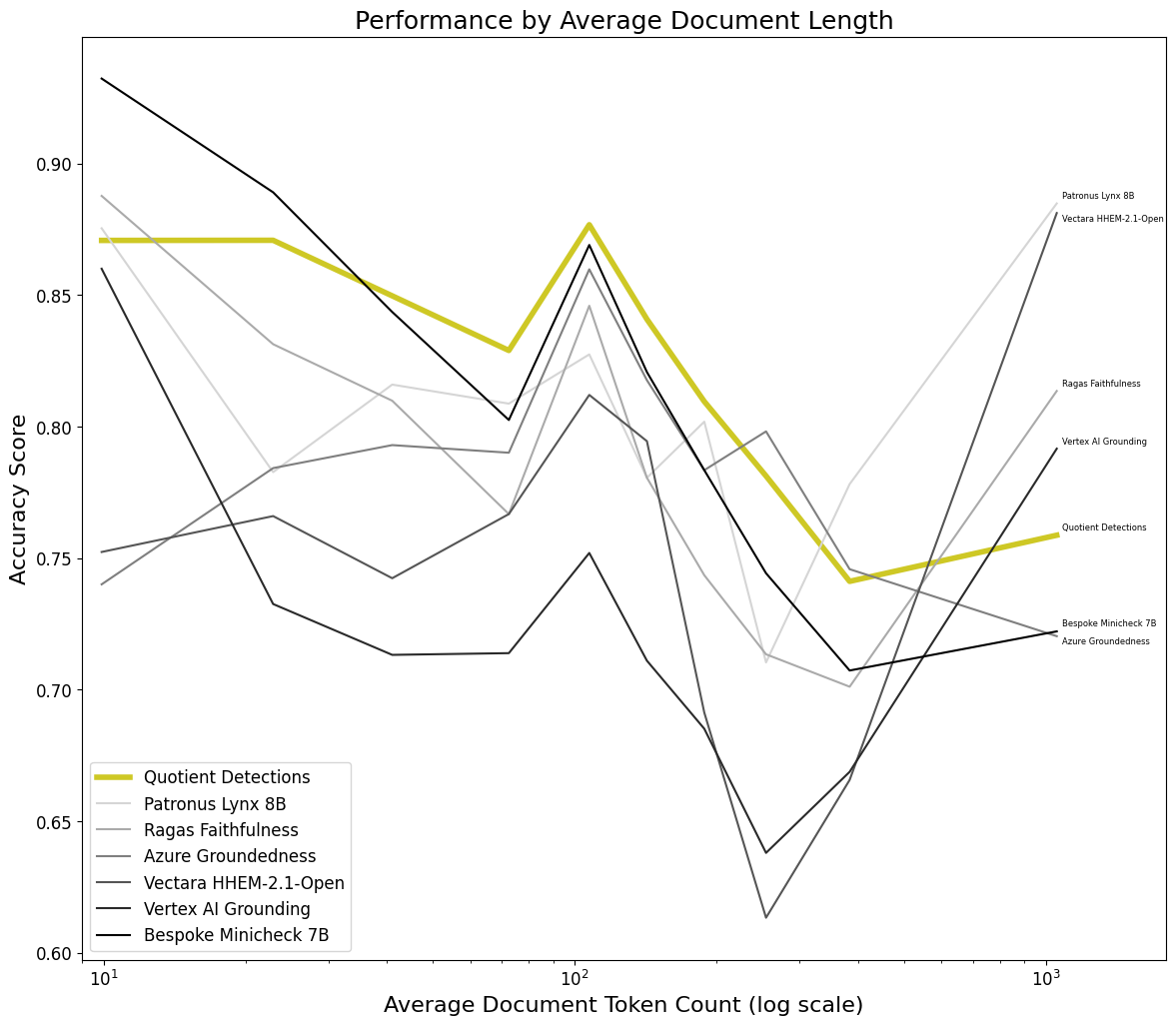}
        \caption{Accuracy by average document token count for each hallucination detection method.}
        \label{fig:pef-doc-length}
    \end{subfigure}%
    ~
    \begin{subfigure}{0.5\textwidth}
        \centering
        \includegraphics[width=\linewidth]{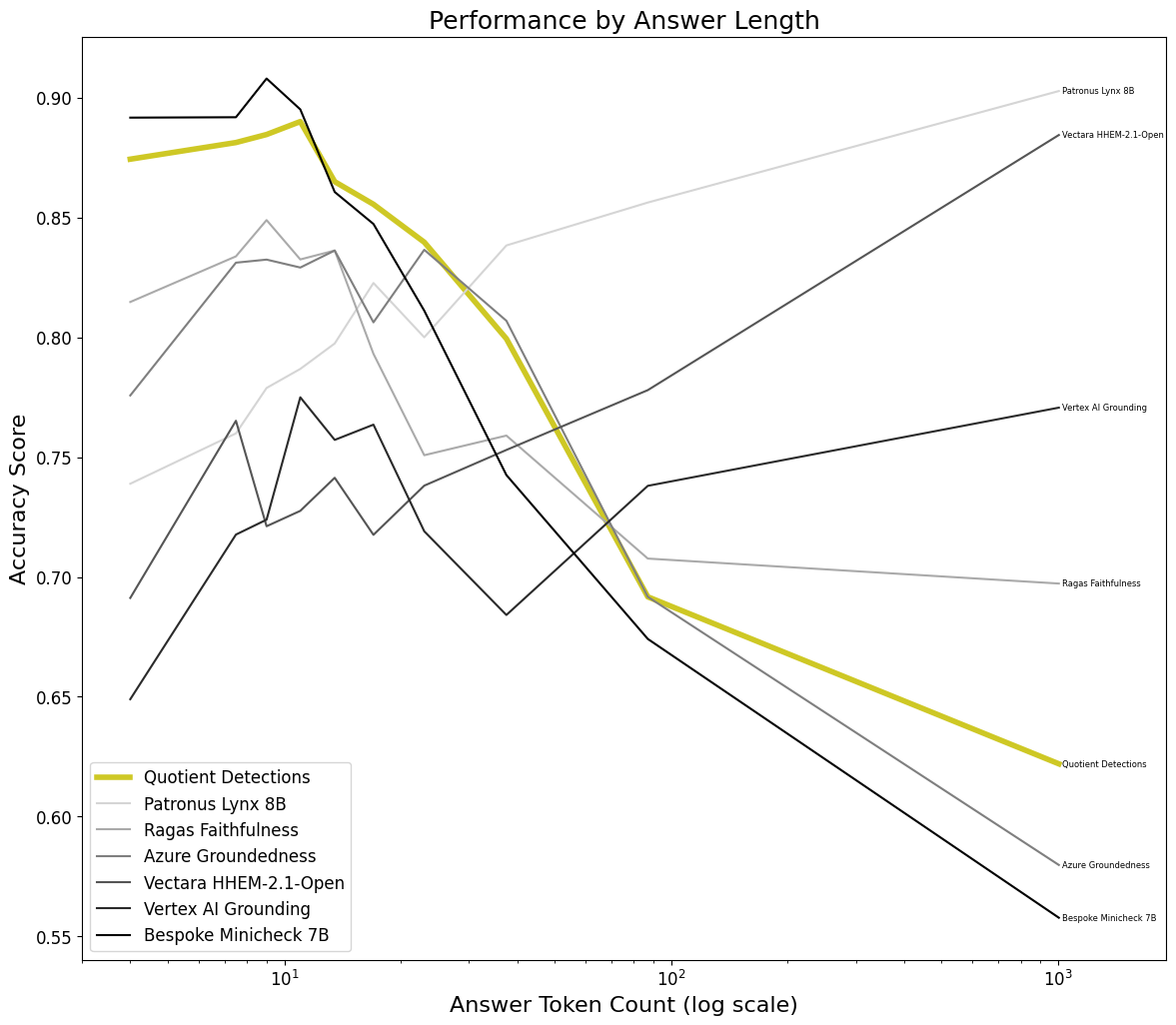}
        \caption{Accuracy by answer token count for each hallucination detection method.}
        \label{fig:perf-ans-length}
    \end{subfigure}
    \caption{Performance of hallucination detection methods as a function of content length. Panel (a) shows accuracy versus average document token count, revealing how different methods handle increasingly complex contexts. Panel (b) shows accuracy versus answer token count, demonstrating performance on longer responses. Both plots show distinct performance patterns: some methods maintain consistent accuracy across lengths while others show clear degradation with longer content.}
    \label{fig:performance-by-length}
\end{figure*}

Figure \ref{fig:performance-by-length} reveals an important pattern in how different architectural approaches handle content length. \textit{Vectara HHEM-2.1-Open} and \textit{Patronus Lynx 8B}—both fine-tuned models that process continuous rather than chunked context—consistently demonstrate superior performance on longer content but struggle with shorter examples. In contrast, sentence-based approaches like \textit{Quotient Detections} and \textit{Bespoke-Minicheck-7B} excel with shorter content ($\sim$200 tokens) typical of NLI and QA tasks, but show degraded performance on long-form summarization examples.

This divergence in performance highlights fundamental trade-offs in context representation for hallucination detection. Continuous-context methods may better preserve document coherence, maintaining critical discourse signals and cross-sentence dependencies that support accurate faithfulness assessment in longer texts. However, sentence-based approaches offer greater precision for granular claim verification in shorter contexts but suffer from information loss when processing longer documents. 

Both sentence-based methods achieve recall values approaching 1.0 on summarization examples, indicating they tend to over-predict hallucinations when evaluating long-form content—likely because sentence isolation disrupts coreference chains and other cross-sentence contextual signals essential for accurate assessment.

These findings suggest several potential improvements for sentence-based detectors. Incorporating sliding window contexts that include neighboring sentences during evaluation could help preserve local coherence. Alternatively, a hierarchical verification approach might first evaluate individual sentences and then perform a second-pass verification using full paragraph context. Such approaches could maintain the granularity advantages of sentence-level detection while addressing the context fragmentation issue.

\subsection{Toward Robust Hallucination Detection}

Future work for \textit{Quotient Detections} will focus on improving performance in long-context scenarios. This includes exploring hybrid approaches that maintain sentence-level granularity while leveraging global document context. The goal is to develop a detection system that remains effective across the full spectrum of generation lengths, from brief factual claims to multi-paragraph summaries.

Overall, our findings emphasize the need for hallucination detection systems that can operate reliably across varied input types, document lengths, and generation formats. HalluMix provides a robust foundation for this line of research, and our evaluation surfaces key directions for future improvement in both model architecture and input handling strategies.

\section{Conclusion}

In this work, we introduced \textit{HalluMix}, a large-scale, task-diverse, and domain-spanning benchmark dataset for evaluating hallucination detection in realistic language generation settings. Unlike prior benchmarks, our dataset reflects the challenges of multi-document grounding and long-form responses common in modern LLM deployments. We systematically evaluated seven detection methods and revealed significant variation in their effectiveness based on input format, content length, and underlying task. Overall, \textit{Quotient Detections} achieves the best performance in accuracy and F1.

Our analysis surfaced several key findings. Some detection models appear to overfit to known datasets, raising concerns about generalization. Sentence-level approaches excel in shorter-content detection but struggle with longer contexts, likely due to loss of inter-sentential coherence. Meanwhile, models evaluated on full continuous text may benefit from preserved context, highlighting the limitations of segmented inputs typical in retrieval-augmented generation pipelines. This finding highlights critical trade-offs between granular claim verification and document-level coherence assessment.

These findings have important implications for LLM deployment in production systems. Organizations implementing RAG-based solutions should carefully consider the limitations of existing hallucination detectors, particularly when working with domain-specific content or long-form outputs. 

Future directions include improving hallucination detection robustness across content lengths, better modeling of discourse-level dependencies, and adapting detectors to handle real-world, chunked input more effectively. We make our benchmark publicly available to facilitate continued research in this critical area of LLM safety and reliability.

\bibliography{references}

\clearpage
\appendix

\section{Appendix}

\begin{table}[ht]
\centering
\begin{tabular}{p{0.2\textwidth}|>{\raggedright\arraybackslash}p{0.7\textwidth}}
\hline
%%%
\textbf{Documents} & 
\begin{itemize}
  \item Due to the Steelers' loss to the Ravens the previous day, the Bengals entered the game as the AFC North champions. The Bengals rushed out to a 14-0 lead in the first half on a McCarron touchdown pass and a Mohamed Sanu rush, but Denver cut the deficit to 11 points as Brandon McManus nailed a short 23-yard field goal with just 18 seconds remaining before halftime. In the second half, momentum shifted mightily after a missed field goal by Mike Nugent in the third. Emmanuel Sanders hauled in an 8-yard pass from Brock Osweiler to cut the deficit to 14-10, and Denver claimed the lead for the first time in the game on a 39-yard touchdown run by C.J. Anderson with 11:17 remaining in the 4th Quarter. The Bengals marched down the field to tie the game on Mike Nugent's season-long 52-yard field goal, making the score 17-17 at the end of regulation. The tired Bengals failed to put any  points on the board in the extra period, allowing a 37-yard McManus field goal to make the score 20-17 Denver. A botched snap on the ensuing Bengals drive was recovered by the Broncos, ending the game and Cincinnati's hopes for a first-round bye in the playoffs. With the loss, the Bengals fell to 11-4 on the season. The loss was also the 10th straight in Denver for the Bengals, dating back to 1975.
\end{itemize} \\
\hline
%%%
 & \\
\textbf{Response} & The first field goal was by the Ravens.\\
 & \\
\hline
 & \\
\textbf{Label} & Hallucinated \\
 & \\
\hline
\end{tabular}
\centering
\caption{\mbox{An example of a hallucinated datapoint in the HalluMix Benchmark.}}
\label{tab:data-example-hallucinated}
\end{table}

\clearpage

\begin{table}[ht]
\centering
\begin{tabular}{p{0.2\textwidth}|>{\raggedright\arraybackslash}p{0.7\textwidth}}
\hline
%%%
\textbf{Documents} & 
\begin{itemize}
  \item Final Fantasy is a Japanese science fantasy anthology media franchise created by Hironobu Sakaguchi and developed and owned by Square Enix (formerly Square).
  \item Peter Wright, a law supervisor for the DNR, told WLUC-TV that the officer was just doing his job. He said the officer believed it was a feral pig, since it had no identifying marks to distinguish him as a pet. 'I want to make it very clear that it's never ever, ever the department's position that we want to shoot people's pets,' said Wright. 'If he had any inkling it was a pet, he absolutely wouldn't have shot it.' Upsetting: The family are now trying to get Caesar's body in order to bury him, but have been told they can only take possession of his ashes . Brandy Savelle and Tony Gervasi are now trying to get Caesar's body back. However they have been told they can only take possession of ashes. Ms Savelle is demanding that some sort of recourse comes out of the situation. 'If it was that big of a mistake then we would like to see better training,' she said. 'Let's learn to identify not just pigs, but all pets.'
  \item God Hates Us All is the eighth studio album by American thrash metal band Slayer .
  \item that's right that's exactly right so but a lot of more women are starting their own businesses i've noticed than
  \item The franchise centers on a series of fantasy and science fantasy role-playing video games. The first game in the series was released in 1987, with 15 numbered main entries having been released to date.
  \item Shortly after 3600 BC Egyptian society began to grow and advance rapidly toward refined civilization .
  \item boy pushing wagon with two pumpkins in it
\end{itemize} \\
\hline
%%%
 & \\
\textbf{Response} & Final Fantasy was created by Hironobu Sakaguchi\\
 & \\
\hline
 & \\
\textbf{Label} & Faithful \\
 & \\
\hline
\end{tabular}
\caption{\mbox{An example of a faithful datapoint in the HalluMix Benchmark.}}
\label{tab:data-example-faithful}
\end{table}

\end{document}